# Using Tabu Search Algorithm for Map Generation in the Terra Mystica Tabletop Game

Alexandr Grichshenko, Luiz Jonata Pires de Araujo, Susanna Gimaeva and Joseph Alexander Brown

*Abstract*— Tabu Search (TS) metaheuristic improves simple local search algorithms (e.g. steepest ascend hill-climbing) by enabling the algorithm to escape local optima points. It has shown to be useful for addressing several combinatorial optimization problems. This paper investigates the performance of TS and considers the effects of the size of the Tabu list and the size of the neighbourhood for a procedural content generation, specifically the generation of maps for a popular tabletop game called Terra Mystica. The results validate the feasibility of the proposed method and how it can be used to generate maps that improve existing maps for the game.

*Index Terms*—Optimization, Procedural Content Generation, Tabu Search, Terra Mystica.

## I. INTRODUCTION

Tabu Search (TS) is a general-purpose metaheuristic which has been used for solving complex combinatorial problems, typically instances of NP-hard problems [1]. TS improves local search algorithms in a sense it contains mechanisms for escaping local optima by maintaining a list of prohibited features, which prevents the algorithm from trying previously considered solutions repeatedly. The performance of TS is supposed to be affected by some parameters like the size of the Tabu list and, like in other metaheuristics, the neighbourhood size [2].

One of the applications of metaheuristics is automatic landscape generation, which is one of the many challenges addressed by designers of online and tabletop games [3]. The reasons for such include complex domain-related requirements when generating realistic maps and the concern of providing a balanced distribution of resources and allow equal victory conditions for players with comparable skills [4].

This is the case in Terra Mystica TM, which is a Euro tabletop game in which players transform the landscape to spread their factions. The distribution of the terrain types across the map affects the cost for such an action and, consequently, the chances of victory for each player. The presence of a complex set of domain-related constraints increases the difficulty of solving this problem.

This study focuses on the performance of TS and the effects of the Tabu list and neighbourhood size when generating maps for TM. In other words, how the quality and the balance, as captured by measurable requirements, change according to these parameters. There has been no practical investigation in this regard in the literature, to the best of the authors' knowledge.

The remaining part of the paper is organized as follows: Section 2 reviews the use of TS for content generation and considerations about its implementation, including the neighbourhood. In section 3, we describe the implementation of the TS algorithm. Next, Section 4 provides an overview of the obtained results and Section 5 elaborates on the effects of different neighbourhood and Tabu list sizes. Lastly, Section 6 summarizes contributions of this work and gives suggestions for future research.

## II. RELATED WORK

Tabu search, which was introduced [1] by Fred W. Glover in 1986, is a metaheuristic that employs local search methods used for mathematical optimization and expands its solution space beyond local optimality. TS enhances local search methods by loosening their rule [5]: if no better solutions are available in the search space at the current step, worsening moves can be accepted. In addition, in order to discourage the search from making cycles, Tabu list of previously visited solutions is maintained. These modifications help the algorithm avoid being stuck in a local optimum.

Several aspects have been noticed to affect Tabu search performance. The size of the Tabu list influences the amount of previously visited neighbours kept in memory. According to Taillard [6], if the Tabu list size is too small, the search process may frequently return to the solutions previously visited, creating cycles. On the other hand, beneficial moves may be rejected with large Tabu list size, leading to the exploration of less optimal solutions and taking a larger number of iterations to converge. One more aspect to consider in TS, likewise in other metaheuristics, is the size of the neighbourhood. Examining the whole neighbourhood in TS generally yields high-quality solutions but can be very computationally expensive if the neighbourhood size is considerable [2].

Map generation is an essential part of automatic content generation in games and offers ground for the application of several metaheuristics [4]. One of the first examples of map generation was reported for Almansur Battlegrounds [7], a turn-based computer game. A similar method was also employed to Dune 2, which is a real-time strategy game in which the terrain affects the performance of the player [8]. Both studies succeeded in producing playable maps, although the authors recognize that map balance issues were not completely resolved. Studies applying genetic algorithms (GA) and genetic programming (GP) for Siphon [9] and Planet Wars [10] have been reported and achieved improve studies rated in a comparatively small search space, which is not the case for more realistic maps. Interestingly, there has been a limited amount of work demonstrating the use of computation techniques for generating maps in tabletop games [11].

An important factor to consider in map generation is providing equal chances of winning to players of equal skills and that no starting position can guarantee victory [12]. According to [3], map balance is often achieved by evaluation functions consisting of features and corresponding weights inferred by the game designer. Another method involves the use of artificial agents to playtest several maps, which is a time-consuming process. Examples of evaluation function-based methods for map balance include studies on StarCraft [13], Civilization [14] and PacMan [15] games. A similar methodology was proposed by Ashlock et.al using dynamic programming to design more suitable fitness functions [16]. Again, there has been no significant amount of work applied to tabletop games, and none in the case of TM.

### III. METHODOLOGY

In this section, we describe the evaluation function based upon a set of requirements to ensure balance in TM maps. Moreover, it presents the implementation details of the experimented TS implementation.

#### A. Map evaluation and search space in Terra Mystica

Maps in TM should attempt to satisfy certain requirements involving the distribution of terrains across the map. The requirements chosen for this study are as follows: a non-river hexagon cannot have one or more neighbours of identical terrain type (REQ1); a river hexagon should have from one to three river neighbours to avoid formation of lakes, which hinder the development of certain factions, e.g. Mermaids (REQ2); all river hexagons should be connected (REQ3); a land hexagon should be a neighbour to at least one other land hexagon that can be terraformed using exactly one spade (REQ4). Terraforming is one of the main actions that a player takes to expand his/her faction, and its cost is directly proportional to the number of spades between the target terrain and the home terrain. For example, two spades separate the mountain (grey) terrain and desert (yellow) terrain shown in Fig 1.

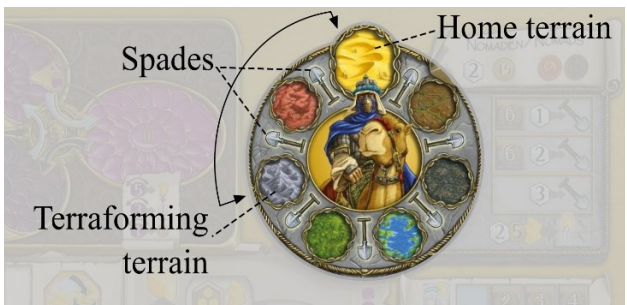

Fig. 1. A player board indicating the terraforming circle.

While REQ1 and REQ3 encourage the generation of maps closely resembling the official map published by the developers, REQ2 and REQ4 prevent the algorithm from introducing imbalanced regions across the map. The evaluation function calculates the number of violations of each requirement and computes the score of the map as the total amount of requirement violations. The objective of the TS is, therefore, to minimize such an objective function.

A TM map contains 77 land tiles (11 of each terrain type) and 36 river tiles, organized in a fixed layout of 9 rows and up to 13 columns. Such characteristics give rise to a search space of approximately $3.7 * 10^{89}$ candidate solutions.

#### B. Tabu Search implementation in this study

Our TS implementation follows the original algorithm introduced by Glover [12] with some domain-specific customization where it is necessary.

Firstly, a candidate solution is an instance of a full map and described as an array of 113 positions containing a char value specifying the terrain type. Second, the stopping condition for TS is met when the running time reaches five minutes, which is an upper bound of convergence time as determined by initial tests. The algorithm receives the following two input parameters: the size of the Tabu list and the size of the neighbourhood in which TS investigates candidate solutions.

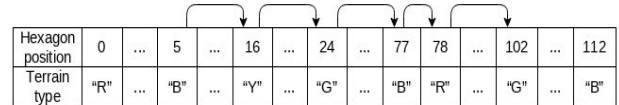

(a) Functionality of a perturb operator

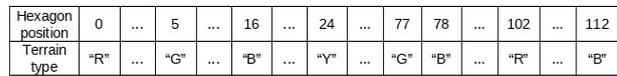

(b) State of the map after perturb operator changes the terrain types

Fig. 2. Perturb operator.

A candidate solution is obtained from the current solution by applying the perturb operator, i.e. changing the terrain types of 6 randomly selected hexagons, as illustrated in Fig. 2. After obtaining the array A of random hexagons, for i in [0; 4], the terrain at position A[i] replaces the terrain at position A[i+1] with A[5] subsequently replacing A[0].

### IV. RESULTS

In this study, the null hypothesis is stated as there is no such correlation between the score of the best solution found by the TS algorithm and the size of the neighbourhood and the size of the Tabu list. Several parameters values have been tested: 10, 20, 30, 50, 75 and 100 were considered for the Tabu list size; 10, 25, 50, 75 and 100 for neighbourhood size. For each hyperparameter, the algorithm was executed 30 times, while the stopping criterion was met when TS running time exceeded five-minutes. Fig. 3 presents the mean score and the standard deviation for each combination of parameters.

The Tabu list size of 50 and neighbourhood size of 75 showed the best mean score (28.3). It can be observed that the performance slightly improves with the increase of the neighbourhood size. However, the size of the Tabu list does not seem to have a significant impact, since the mean scores across all values of this parameter are similar. The map with the best overall score shown in Fig. 4 was obtained with the Tabu list size of 75 and neighbourhood size of 75, with its score equal to 14.0.

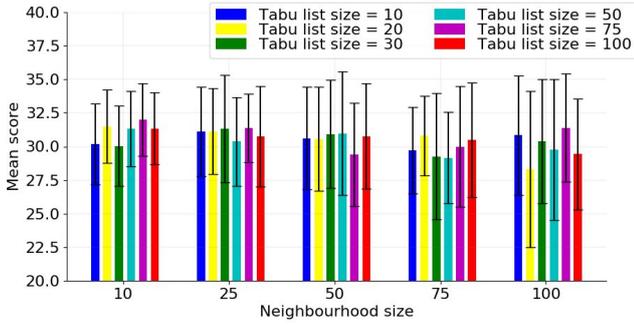

Fig. 3. Mean score for each neighbourhood size.

After obtaining the results of hyperparameter optimization, the one-way ANOVA (ANalysis Of Variance) [13] was performed to determine whether the dependence of evaluation score on neighbourhood size and tabu list size are statistically significant. All characteristics of the data that are relevant for ANOVA are detailed in Table 1. For determining the critical value, alpha = 0.05 was used, which corresponds to a 95% confidence level.

The results do not allow one to reject the null hypothesis and state that this parameter has a statistically significant impact on the results. Conversely, F-value for neighbourhood size is larger than the critical value, which suggests that this parameter can affect the results of the TS algorithm.

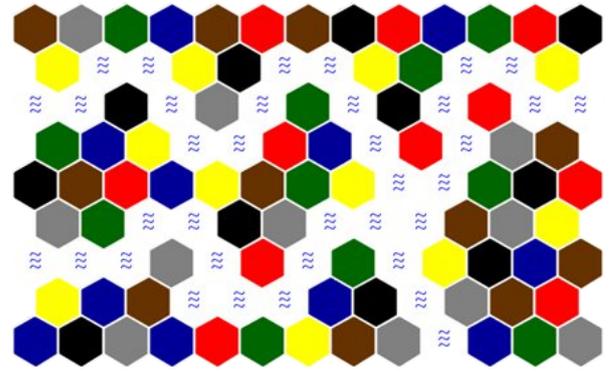

(c) FireIce

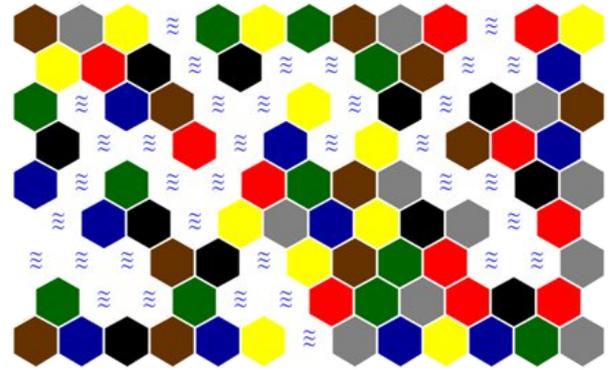

(d) Best map obtained by TS (Tabu list size 75 and neigborhood size 75).

Fig. 4. Terra Mystica maps.

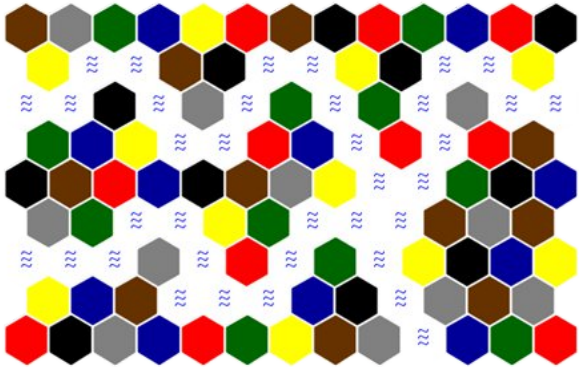

(a) Original Map

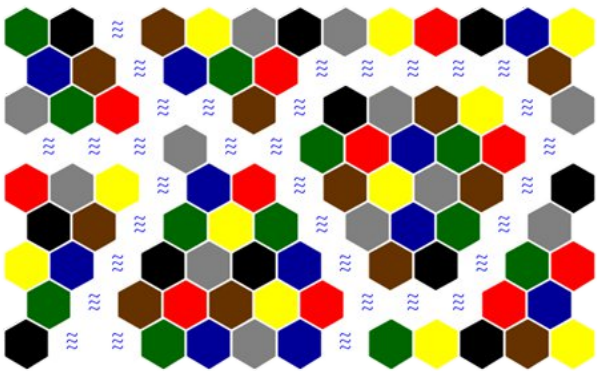

(b) Fjords

TABLE I: ONE-WAY ANOVA RESULTS

| Parameter | $SSO_b$ | $SSO_w$ | $df_1$ | $df_2$ | F-value | F-min |
|---|---|---|---|---|---|---|
| Size of the neighbourhood | 209.8 | 13597 | 4 | 895 | 3.45 | 2.39 |
| Tabu list size | 23.84 | 13783 | 5 | 894 | 0.31 | 2.22 |

## V. DISCUSSION

The results show that the Tabu list size in TS did not significantly change the performance of the algorithm for this particular problem domain. This contrasts with several other applications, where the TS performance is typically affected by this parameter [14,15]. Neighbourhood size, on the other hand, influences the results directly, which is consistent with some of the previously performed experiments reporting a positive correlation between the values [16].

This study has validated the use of TS for search-based procedural content generation, in particular, map generation for Terra Mystica tabletop game. The requirements for a balanced resource distribution among the factions in the game have been met. This fact challenges the doubts on the use of the algorithms for creative tasks because the utilized technique, as well as potentially many others, can enliven the game by introducing new maps and, as a consequence, richer playing experience. Interestingly, some requirements are often more disregarded than others, as shown in Table 2.

Table 2 shows that the maps released by the developers do not violate requirements REQ1 and REQ3, while maps generated by Tabu search perform better with REQ2 and

REQ3. The success of REQ3 can be explained by the relative simplicity of its implementation using the DFS graph traversal algorithm to find strongly connected components (i.e. rivers). Issues regarding REQ4 are present in all maps, as this requirement is violated the most. The reason for this might be the nature of the perturb operator we have used for this study. More sophisticated operators could be used to refine the performance of the Tabu search on REQ4.

TABLE II: The number of times each requirement is violated

| Map | Score ($F_{tot}$) | REQ1 | REQ2 | REQ3 | REQ4 |
|---|---|---|---|---|---|
| Tabu Search | 14 | 4 | 3 | 1 | 6 |
| Original Map | 9 | 0 | 3 | 0 | 6 |
| Fire and Ice | 12 | 0 | 3 | 0 | 9 |
| Fjords | 9 | 0 | 0 | 0 | 9 |

The need for new experiences in TM board game has driven the PCG using metaheuristics to be applied for balanced maps generation. User satisfaction from the engaging game with fair faction distribution in the map can be achieved with the aforementioned algorithm. The success of the method can be seen in online platforms like `TM AI' https://lodev.org/tmai. Taking into account the number of released maps, only two so far, this algorithm can produce the maps which can potentially be commercialized.

## VI. Conclusion

The paper demonstrates how Tabu search metaheuristic can be applied to the task of generating maps for the Terra Mystica board game. Solutions for the tackled problem were highly anticipated by the large community of players and developers. Subsequently, the study showed how the generate-and-test approach could be used to obtain balanced and playable maps that can greatly enhance the replayability value of the game. Moreover, a statistical analysis of the effect of neighbourhood and Tabu list size in TS on the evaluation score was conducted. ANOVA failed to prove or disprove the hypothesis of direct dependency between Tabu list size and results of TS. However, statistical analysis of neighbourhood size suggested that there exists a correlation between the parameter and evaluation score.

In future research, the focus should be placed on the expansion of the set of requirements and evaluation function with the use of mathematical modelling. It is also significant to apply other combinatorial optimization techniques including Particle Swarm and Ant Colony Optimization to fully estimate the potential of such evolutionary algorithms with regard to PCG. Moreover, the study on the effects of neighbourhood size needs to be investigated with regard to a larger set of metaheuristics and problem domains.

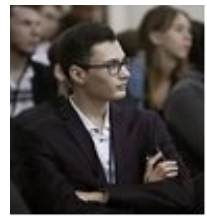
**Alexandr Grichshenko** is a second-year bachelor student of Computer Science at Innopolis University, Russia. Conducted research, published several academic papers and participated in internal projects in the lab of Machine Learning and Artificial Intelligence. His research interests include optimization, machine learning, algorithm selection and mathematical programming applied to real-world problems.

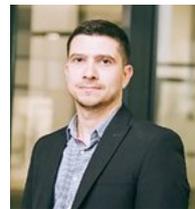
**Luiz Jonata Pires de Araujo** is a PhD in Computer Science at the University of Nottingham (UK), having integrated the Automated Scheduling, optimisAtion and Planning (ASAP) research group.
Currently, he works as a Faculty Fellow at Innopolis University, Russian Federation, in the lab of Machine Learning and Artificial Intelligence. His areas of interest include optimization, evolutionary algorithms, cutting and packing, and 3D Printing. Recently assumed the position of co-leader of the internal project in Innopolis working on BioDynaMo framework, a simulation tool maintained by CERN.

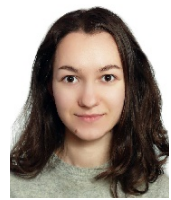
**Susanna Gimaeva** is a third-year bachelor student of Computer Science at Innopolis University, Russia. Susanna has conducted research in the lab of Software Engineering as well as Machine Learning and

Artificial Intelligence, published several academic papers and presented her work in international conferences. Her research interests include optimization and machine learning.

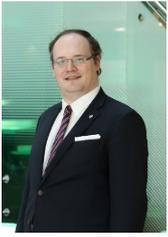

**Joseph Alexander Brown** received the B.Sc. (Hons.) with first class standing in computer science with a concentration in software engineering, and M.Sc. in computer science from Brock University, St. Catharines, ON, Canada in 2007 and 2009, respectively. He received the Ph.D. in computer science from the University of Guelph in 2014.
He previously worked for Magna International Inc. as a Manufacturing Systems Analyst and as a visiting researcher at ITU Copenhagen in their Games Group.

He is currently an Assistant Professor and Head of the Artificial Intelligence in Games Development Lab at Innopolis University in Innopolis, Republic of Tatarstan, Russia and an Adjunct Professor of Computer Science at Brock University, St. Catharines, ON, Canada.

Dr. Brown is a Senior Member of the IEEE, a chair of the yearly Procedural Content Generation Jam (ProcJam), the proceedings chair for the IEEE 2013 Conference on Computational Intelligence in Games (CIG) and is Vice Chair for the IEEE Committee on Games.